\documentclass{article}

\usepackage{graphicx}
\usepackage{amsmath}

 \usepackage[preprint]{neurips_2025}

\usepackage[utf8]{inputenc} 
\usepackage[T1]{fontenc}    
\usepackage{hyperref}       
\usepackage{url}            
\usepackage{booktabs}       
\usepackage{amsfonts}       
\usepackage{nicefrac}       
\usepackage{microtype}      
\usepackage{xcolor}         
\usepackage{natbib}
\usepackage{bibentry}

\title{Physical Consistency of Aurora's Encoder: A Quantitative Study}

\author{%
  Benjamin Richards \\
  IMT Atlantique \\
  Nantes, France \\
  \texttt{benjamin.richards@imt-atlantique.net} \\
  \And
  Pushpa Kumar Balan \\
  University of Central Missouri \\
  Warrensburg, MO, USA \\
  \texttt{pushpakumarbalan@gmail.com} \\
}

\newcommand{\LandSeaAccuracy}{99.89}
\newcommand{\LandSeaPrecision}{99.73}
\newcommand{\LandSeaRecall}{99.76}

\newcommand{\ExtremeHeatLowAccuracy}{95.71}   
\newcommand{\ExtremeHeatMediumAccuracy}{97.19} 
\newcommand{\ExtremeHeatHighAccuracy}{97.76}  
\newcommand{\ExtremeHeatExtremeAccuracy}{98.99} 

\newcommand{\ExtremeHeatLowPrecision}{96.28}
\newcommand{\ExtremeHeatMediumPrecision}{93.50}
\newcommand{\ExtremeHeatHighPrecision}{89.76}
\newcommand{\ExtremeHeatExtremePrecision}{89.19}

\newcommand{\ExtremeHeatLowRecall}{95.12}
\newcommand{\ExtremeHeatMediumRecall}{91.52}
\newcommand{\ExtremeHeatHighRecall}{88.37}
\newcommand{\ExtremeHeatExtremeRecall}{79.51}

\newcommand{\AtmosInstabilityLowAccuracy}{98.83}   
\newcommand{\AtmosInstabilityHighAccuracy}{99.27}  

\newcommand{\AtmosInstabilityLowPrecision}{98.31}
\newcommand{\AtmosInstabilityHighPrecision}{93.60}

\newcommand{\AtmosInstabilityLowRecall}{98.18}
\newcommand{\AtmosInstabilityHighRecall}{92.07}

\newcommand{\LandSeaConceptProbCorr}{98.06}
\newcommand{\LandSeaMeanConceptZero}{-3.13}
\newcommand{\LandSeaMeanConceptOne}{2.73}
\newcommand{\LandSeaMeanConceptSep}{5.87}

\newcommand{\ExtremeHeatLowConceptProbCorr}{77.88}
\newcommand{\ExtremeHeatLowConceptZero}{-0.95}
\newcommand{\ExtremeHeatLowConceptOne}{0.64}
\newcommand{\ExtremeHeatLowConceptSep}{1.58}

\newcommand{\ExtremeHeatMediumConceptProbCorr}{72.44}
\newcommand{\ExtremeHeatMediumConceptZero}{-1.31}
\newcommand{\ExtremeHeatMediumConceptOne}{0.79}
\newcommand{\ExtremeHeatMediumConceptSep}{2.09}

\newcommand{\ExtremeHeatHighConceptProbCorr}{74.88}
\newcommand{\ExtremeHeatHighConceptZero}{-2.00}
\newcommand{\ExtremeHeatHighConceptOne}{1.07}
\newcommand{\ExtremeHeatHighConceptSep}{3.07}

\newcommand{\ExtremeHeatExtremeConceptProbCorr}{80.81}
\newcommand{\ExtremeHeatExtremeConceptZero}{-3.55}
\newcommand{\ExtremeHeatExtremeConceptOne}{1.97}
\newcommand{\ExtremeHeatExtremeConceptSep}{5.51}

\newcommand{\AtmosInstabilityLowConceptProbCorr}{80.28}
\newcommand{\AtmosInstabilityLowConceptZero}{-1.22}
\newcommand{\AtmosInstabilityLowConceptOne}{0.71}
\newcommand{\AtmosInstabilityLowConceptSep}{1.94}

\newcommand{\AtmosInstabilityHighConceptProbCorr}{44.62}
\newcommand{\AtmosInstabilityHighConceptZero}{-1.67}
\newcommand{\AtmosInstabilityHighConceptOne}{0.28}
\newcommand{\AtmosInstabilityHighConceptSep}{1.95}

\begin{document}

\maketitle

\begin{abstract}
    The high accuracy of large-scale weather forecasting models like Aurora is often accompanied by a lack of transparency, as their internal representations remain largely opaque. This "black box" nature hinders their adoption in high-stakes operational settings. In this work, we probe the physical consistency of Aurora's encoder by investigating whether its latent representations align with known physical and meteorological concepts. Using a large-scale dataset of embeddings, we train linear classifiers to identify three distinct concepts: the fundamental land-sea boundary, high-impact extreme temperature events, and atmospheric instability. Our findings provide quantitative evidence that Aurora learns physically consistent features, while also highlighting its limitations in capturing the rarest events. This work underscores the critical need for interpretability methods to validate and build trust in the next generation of AI-driven weather models. \par The code is available at \href{https://github.com/richardsbenjamin/auroraencoderanalysis}{https://github.com/richardsbenjamin/auroraencoderanalysis}.
\end{abstract}

\section{Introduction}
Global weather forecasting, once dominated by physics-based Numerical Weather Prediction (NWP) models, is rapidly shifting toward data-driven deep learning systems. Recent models such as Aurora, a transformer-based foundation model, achieve state-of-the-art accuracy at far lower computational cost \cite{Aurora}. However, their high performance comes with limited transparency. The internal latent representations of such models remain opaque, posing challenges for trust and reliability in operational forecasting.

This study extends earlier internal analyses to test the physical consistency of Aurora’s encoder. Using a large dataset, we examine whether its latent representations capture three key meteorological concepts:
\begin{enumerate}
    \item The land–sea boundary
    \item Extreme temperature events
    \item Indicators of atmospheric instability
\end{enumerate}
By assessing the model’s ability to encode these physical relationships, we aim to advance understanding of how foundation models represent atmospheric processes and to inform efforts toward more interpretable AI-based weather prediction.

\section{Data}
The input data for this study are taken from the ERA5 reanalysis dataset \cite{era5}, spanning the period from 2024-07-13T12:00 to 2024-07-16T00:00, with 6-hourly time steps on a 0.25° grid. Each consecutive pair of time points is processed by the Aurora encoder to produce latent representations, yielding a sequence of ten consecutive latent time steps.

The latent representation is divided into two components:
(i) a surface embedding describing near-surface variables (2-m air temperature, 10-m winds, and sea-level pressure) together with static fields (land–sea mask, soil type, and surface geopotential); and
(ii) an atmospheric embedding capturing temperature, winds, humidity, and geopotential height across pressure levels.
While the raw ERA5 data include 13 pressure levels, the latent representation is defined at three latent pressure levels.

Binary concept masks are constructed from several data sources. The land–sea mask is obtained directly from the ERA5 dataset through the Climate Data Store API and is static in time. To identify extreme temperature conditions, we use ECMWF climate projection statistics to define percentile thresholds (75th, 90th, 95th, and 99th) of 2-m temperature across Europe. ERA5 values exceeding these thresholds are labelled as extreme and resampled to match the spatial resolution of the Aurora encoder.

For atmospheric instability, data are prepared for subsequent computation of the K-index, which serves as a proxy for convective potential \cite{kindex}. Temperature and relative-humidity fields from ERA5 are extracted at the relevant pressure levels (850, 700, and 500 hPa) to allow later derivation of the dew-point temperature and K-index values (see Section~\ref{Methodology}).

\section{Methodology}
\label{Methodology}
Our methodology is based on the principle of linear probing, in which simple and interpretable models are trained to decode high-level concepts from fixed, high-dimensional embeddings produced by a pre-trained model. 

\subsection{Derived Variables: Dew Point and K-index}
To quantify atmospheric instability, we compute the K-index \cite{kindex} from ERA5 temperature and humidity data at 850, 700, and 500 hPa pressure levels. Temperature values are provided directly by ERA5, while the dew-point temperature is derived from the relative humidity using the following relations \cite{dewpoint}:
$$
\phi(T, \mathrm{RH}) = \ln\!\left(\frac{\mathrm{RH}}{100}\right) + \frac{a\,T}{b + T}
$$
$$
T_{\text{dew}} = \frac{b\,\phi(T, \mathrm{RH})}{a - \phi(T, \mathrm{RH})}
$$
where $a=17.625$, $b=243.04^\circ$, $T$ is the temperature, and $RH$ is the relative humidity at a given level. 
The computed dew-point values are then used to evaluate the K-index:
$$
K=(T_{850}-T_{500})+T_{d_{850}}-(T_{700}-T_{d_{700}})
$$
which serves as a diagnostic for convective potential, with higher values indicating increased thunderstorm likelihood. 

\subsection{Latent-Space Analysis}
The latent embeddings obtained from the Aurora model are analysed to investigate their internal representation of physical concepts.

\paragraph{Principal Component Analysis (PCA)}
We first apply PCA to the embedded space to reduce dimensionality and explore the structure of the learned representation. The first two components are used for visual inspection, with data points coloured by concept labels (e.g., land–sea, extreme temperature, instability). This step serves as a qualitative assessment of the embedding space and is not used for quantitative evaluation.

\paragraph{Linear Probing}
To test whether specific concepts are linearly separable in the embedding space, we train a logistic regression classifier for each concept (land–sea, extreme heat, and atmospheric instability). The model predicts binary labels associated with each concept using the frozen latent embeddings as input features.

\paragraph{Concept Vector Analysis}
Following the approach of linear concept activation, we derive a concept vector for each classifier by taking the learned coefficient vector of the trained logistic regression model. This vector represents the direction in the embedding space most associated with the given concept. For a set of embeddings $E$ and concept vector $v_c$, we compute the concept scores as:
$$s=E \cdot \frac{v_c}{ \left\| v_c \right\| }$$
In addition to the concept scores, we obtain the predictions from the logistic model by passing the whole latent dataset to obtain the class predictions, as well as the associated probabilities of obtaining a prediction of class 1. We subsequently calculate various metrics based on the concept score. Specifically, we calculate:
\begin{itemize}
    \item Correlation between the concept scores and the corresponding probabilities. 
    \item Mean concept scores for class 0
    \item Mean concept scores for class 1
    \item Concept score separation as the difference between the mean concept scores.
\end{itemize}

\section{Results}
\label{sec:results}

We present the results of our probing experiments, beginning with visualizations from PCA, followed by the quantitative accuracy of our linear probes, and concluding with an analysis of the relationships between the learned concept vectors.

\subsection{PCA Visualization}

\begin{figure}[h]
  \centering
  \includegraphics[width=0.9\linewidth]{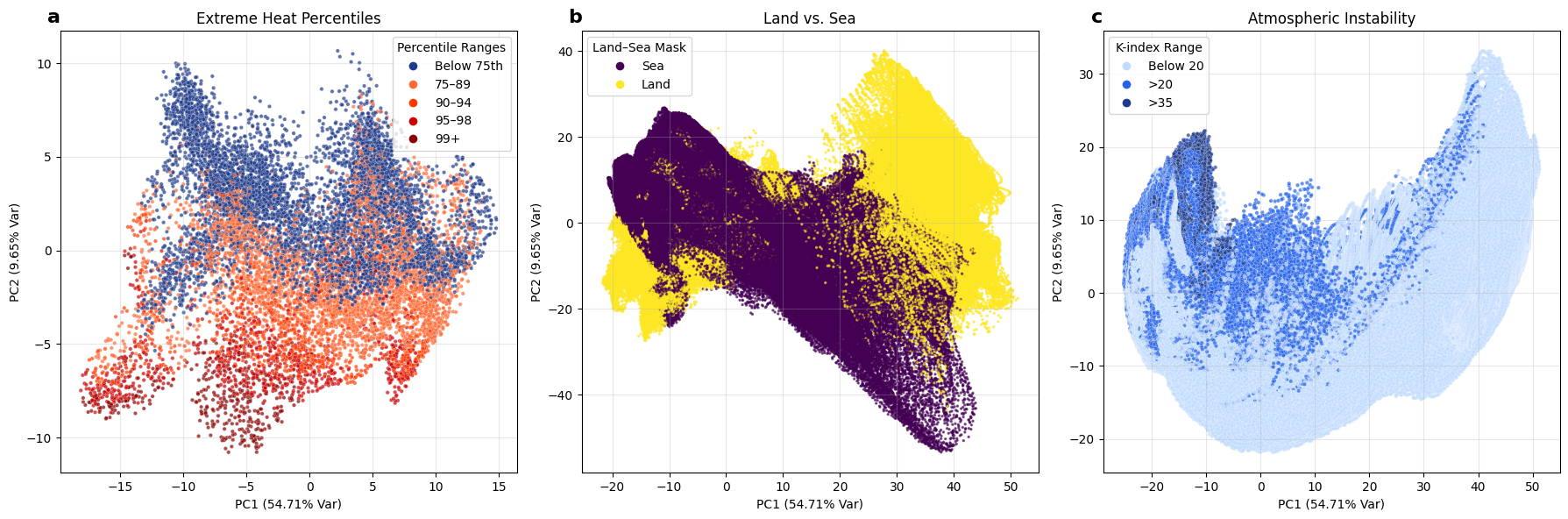}
  \caption{
    PCA projections of the embedding space for the three classification tasks:
    (a) Land–Sea distinction,
    (b) Extreme Heat,
    and (c) Atmospheric Instability.
    Colours indicate class membership or percentile bins, illustrating the separability
    of each concept in the latent space.
  }
  \label{fig:pca_plots}
\end{figure}

To qualitatively assess the structure of the latent space, we applied PCA to the surface and atmospheric embeddings and projected the samples onto their first two principal components. Figure~\ref{fig:pca_plots} shows these projections for our three concepts. For the extreme heat, we see a clear gradient: moderate events are widely dispersed, whereas more severe extremes form a distinct cluster. This indicates that event intensity is represented along a consistent axis of variation within the latent space. For the land-sea distinction, PC1 is clearly separating much of the land and sea data, with land points dominating the right side of the plot and sea points clustering toward the center and left. Overlap in the center suggests that there are regions where land and sea data are not cleanly separable in this latent space (e.g. coastlines). For the atmospheric instability, there is a pronounced gradient in K-index values across the latent space. Most low K-index values dominate the lower and right regions of the plot, while progressively higher values cluster toward the top-left quadrant. The highest K-index values are tightly clustered in the top-left region, suggesting that this area of the latent space captures specific atmospheric states associated with severe instability (e.g., strong convection potential).

\subsection{Linear Probing Accuracy}
Table~\ref{tab:log-reg-metrics} summarizes the classification performance. The Land-Sea Distinction model performs nearly perfectly across all metrics, with accuracy, precision, and recall above 99.7\%. For Extreme Heat, accuracy remains high across percentiles, but recall declines as the threshold becomes more extreme, while precision stays high. This indicates the model is conservative in identifying rare heat events. It correctly labels them when it does, but misses many actual cases. A similar trend is seen in Atmospheric Instability, where higher thresholds show strong precision but reduced recall, suggesting increased caution in predicting more extreme conditions.

\begin{table}[h]
  \caption{Logistic regression classification metrics for each concept.}
  \label{tab:log-reg-metrics}
  \centering
  \begin{tabular}{lccc}
    \toprule
    \textbf{Concept} & \textbf{Accuracy (\%)} & \textbf{Precision (\%)} & \textbf{Recall (\%)} \\
    \midrule
    Land-Sea Distinction & \LandSeaAccuracy & \LandSeaPrecision & \LandSeaRecall \\
    Extreme Heat (75th percentile) & \ExtremeHeatLowAccuracy & \ExtremeHeatLowPrecision & \ExtremeHeatLowRecall \\
    Extreme Heat (90th percentile) & \ExtremeHeatMediumAccuracy & \ExtremeHeatMediumPrecision & \ExtremeHeatMediumRecall \\
    Extreme Heat (95th percentile) & \ExtremeHeatHighAccuracy & \ExtremeHeatHighPrecision & \ExtremeHeatHighRecall \\
    Extreme Heat (99th percentile) & \ExtremeHeatExtremeAccuracy & \ExtremeHeatExtremePrecision & \ExtremeHeatExtremeRecall \\
    Atmospheric Instability (K20) & \AtmosInstabilityLowAccuracy & \AtmosInstabilityLowPrecision & \AtmosInstabilityLowRecall \\
    Atmospheric Instability (K35) & \AtmosInstabilityHighAccuracy & \AtmosInstabilityHighPrecision & \AtmosInstabilityHighRecall \\
    \bottomrule
  \end{tabular}
\end{table}

\subsection{Concept Vector Analysis}

Table~\ref{tab:concept-prob-corr} shows various metrics based on the concept vectors. The land-sea distinction exhibits the strongest alignment with model predictions, showing a high probability correlation (98.06\%) and the largest concept separation (5.87), indicating the model heavily relies on this feature. Extreme heat concepts become more influential at higher percentiles. While the 75th percentile shows moderate correlation (77.88\%) and low separation (1.58), the 99th percentile displays both high correlation (80.81\%) and strong separation (5.51), suggesting the model is more sensitive to severe heat extremes. Atmospheric instability (K20) is moderately aligned with the model (80.81\% correlation, 1.94 separation), while K35 shows weaker correlation (44.62\%) despite similar separation, implying limited influence.

\begin{table}[h]
  \caption{Concept representation metrics.}
  \label{tab:concept-prob-corr}
  \centering
  \begin{tabular}{lcccc}
    \toprule
    \textbf{Concept} & \textbf{Prob. Corr. (\%)} & \textbf{Mean (0)} & \textbf{Mean (1)} & \textbf{Separation} \\
    \midrule
    Land-Sea Distinction & \LandSeaConceptProbCorr & \LandSeaMeanConceptZero & \LandSeaMeanConceptOne & \LandSeaMeanConceptSep \\
    Extreme Heat (75th percentile) & \ExtremeHeatLowConceptProbCorr & \ExtremeHeatLowConceptZero & \ExtremeHeatLowConceptOne & \ExtremeHeatLowConceptSep \\
    Extreme Heat (90th percentile) & \ExtremeHeatMediumConceptProbCorr & \ExtremeHeatMediumConceptZero & \ExtremeHeatMediumConceptOne & \ExtremeHeatMediumConceptSep \\
    Extreme Heat (95th percentile) & \ExtremeHeatHighConceptProbCorr & \ExtremeHeatHighConceptZero & \ExtremeHeatHighConceptOne & \ExtremeHeatHighConceptSep \\
    Extreme Heat (99th percentile) & \ExtremeHeatExtremeConceptProbCorr & \ExtremeHeatExtremeConceptZero & \ExtremeHeatExtremeConceptOne & \ExtremeHeatExtremeConceptSep \\
    Atmospheric Instability (K20) & \AtmosInstabilityLowConceptProbCorr & \AtmosInstabilityLowConceptZero & \AtmosInstabilityLowConceptOne & \AtmosInstabilityLowConceptSep \\
    Atmospheric Instability (K35) & \AtmosInstabilityHighConceptProbCorr & \AtmosInstabilityHighConceptZero & \AtmosInstabilityHighConceptOne & \AtmosInstabilityHighConceptSep \\
    \bottomrule
  \end{tabular}
\end{table}

\section{Conclusion}

Aurora’s latent space captures key physical features with varying fidelity. The land–sea distinction is nearly perfectly represented, while extreme heat and atmospheric instability are encoded more effectively at higher severities, showing high precision but limited recall for rare events. Latent space projections reveal coherent gradients corresponding to event intensity and geographic separation. These results demonstrate that Aurora encodes meaningful meteorological concepts while highlighting limitations in representing rare extremes, emphasizing the need for interpretability in AI-driven weather forecasting.

\bibliographystyle{plainnat}
\bibliography{ref}
{
\small

\end{document}